\documentclass[sigconf,nonacm]{acmart}

\renewcommand\footnotetextcopyrightpermission[1]{} 
\setcopyright{acmlicensed}

\acmConference[E-Energy '25]{Proceedings of the 2025 ACM International Conference on Future Energy Systems}{June 17--20, 2025}{Rotterdam, Netherlands}

\begin{document}

\title{Enhanced Load Forecasting with GAT-LSTM: Leveraging Grid and Temporal Features}

\author{Ugochukwu Orji}
\orcid{0009-0003-9973-8231}
\affiliation{
  \institution{Jheronimus Academy of Data Science (JADS), Tilburg University}
  \city{Tilburg}
  \country{The Netherlands}
}
\email{u.e.orji@tilburguniversity.edu}

\author{\c{C}i\c{c}ek G\"uven}
\orcid{0000-0002-1939-8325}
\affiliation{
  \institution{Department of Cognitive Science and Artificial Intelligence, Tilburg University}
  \city{Tilburg}
  \country{The Netherlands}
}
\email{c.guven@tilburguniversity.edu}

\author{Dan Stowell}
\orcid{0000-0001-8068-3769}
\affiliation{
  \institution{Department of Cognitive Science and Artificial Intelligence, Tilburg University}
  \city{Tilburg}
   \country{The Netherlands}
}
\affiliation{
   \institution{Naturalis Biodiversity Center}
  \city{Leiden}
  \country{The Netherlands}
}

\email{d.stowell@tilburguniversity.edu}

\renewcommand{\shortauthors}{Orji et al.}

\begin{abstract}
Accurate power load forecasting is essential for the efficient operation and planning of electrical grids, particularly given the increased variability and complexity introduced by renewable energy sources. This paper introduces GAT-LSTM, a hybrid model that combines Graph Attention Networks (GAT) and Long Short-Term Memory (LSTM) networks. A key innovation of the model is the incorporation of edge attributes, such as line capacities and efficiencies, into the attention mechanism, enabling it to dynamically capture spatial relationships grounded in grid-specific physical and operational constraints. Additionally, by employing an early fusion of spatial graph embeddings and temporal sequence features, the model effectively learns and predicts complex interactions between spatial dependencies and temporal patterns, providing a realistic representation of the dynamics of power grids. Experimental evaluations on the Brazilian Electricity System dataset demonstrate that the GAT-LSTM model significantly outperforms state-of-the-art models, achieving reductions of 21. 8\% in MAE, 15. 9\% in RMSE and 20. 2\% in MAPE. These results underscore the robustness and adaptability of the GAT-LSTM model, establishing it as a powerful tool for applications in grid management and energy planning.
\end{abstract}


\maketitle

\section{Introduction}

Modern power systems are intricate networks of interconnected components responsible for generating, transmitting, distributing, and utilizing electricity. A key challenge in these systems is maintaining the balance between supply and demand, especially with the rapid integration of renewable energy sources (RES) such as solar and wind into grid operations. The intermittent nature of RES introduces variability, which requires the deployment of new technologies that enhance grid flexibility and enable real-time responses.

Consequently, accurate demand estimation is fundamental for effective power planning as matching supply with demand is essential to maintain grid stability. Note that errors in demand forecasting can result in significant financial costs, with even a 1\% forecasting error potentially leading to hundreds of thousands of dollars in losses per GWh \cite{Mansoor2024}. Since forecasting plays a critical role in numerous tasks, improving both the accuracy and efficiency of forecasting processes is crucial.

Traditionally, load forecasting has relied on statistical methods that estimate relationships within time-series data, sometimes incorporating external factors. Classical statistical models such as ARIMA~\cite{Chodakowska2021} and exponential smoothing~\cite{Lima2019,Taylor2009}, as well as various machine learning approaches~\cite{Bunnoon2010}, have been utilized. In addition, load forecasting is influenced by numerous factors, many of which are location-specific and depend on the equipment used~\cite{Mamun2020}. Factors such as weather conditions, demographics, socioeconomic variables, and special events such as holidays that can impact power demand are often incorporated as input variables in forecast models~\cite{Ahmed2020}.

In modern electricity markets, forecasts are generated for various time horizons, each supporting grid operations and market management. Very short-term forecasts, spanning seconds to minutes, are critical for emergency operations and optimizing Distributed Energy Resources (DER). Short-term forecasts, ranging from minutes to a day ahead, are vital for real-time market trading, power plant operations, grid balancing, and managing operating reserves. Medium-term forecasts, covering days to months, aid in pre-dispatch, unit commitment, and maintenance planning, while long-term forecasts, extending months to years, support system planning, investment decisions, and maintenance scheduling~\cite{Pinson2013, Zavadil2013, Dannecker2015}.

Among these, short-term load forecasting (STLF) is particularly crucial for daily balancing in grid operations and facilitates real-time decision-making for unit dispatching, peak load analysis, and automatic generation control, ensuring efficient grid operation under dynamic conditions, especially as RES are integrated~\cite{Zavadil2013, Dannecker2015, Kyriakides2007,Venayagamoorthy2012}.

Artificial intelligence (AI) techniques and deep neural networks (DNNs) have gained popularity for load forecasting, including expert systems~\cite{Rahman1988}, support vector machines (SVM)~\cite{Dong2022}, fuzzy logic~\cite{Pandian2006}, artificial neural networks (ANNs)~\cite{Kouhi2013} and long-short-term memory networks (LSTMs)~\cite{Jin2022}. However, despite their ability to incorporate external influences, these models have limitations, such as the risk of getting stuck in local minima, overfitting~\cite{Zhang2018}, and inability to fully capture complex spatial-temporal dependencies.

To address these challenges, hybrid models that combine multiple forecasting methods have been proposed to further enhance accuracy and reliability~\cite{Mamun2020,Eandi2022,Lin2021}. These hybrid models offer the advantage of capturing the spatial and temporal features of the electricity load while addressing the limitations of individual methods. However, despite their advantages, hybrid models also introduce additional complexity in model design, implementation, and parameter optimization~\cite{Mamun2020}.

\subsection*{Key Challenges with Traditional and State-of-the-Art Load Forecasting Models}
\begin{enumerate}
    \item \textbf{Limited Incorporation of Grid-Specific Features:} Traditional load forecasting models, and even the current advanced methods, often overlook or inadequately handle grid-specific information (e.g. transmission capacities, efficiencies), which are essential for capturing power flow constraints and line losses in the grid. These models focus mainly on dynamic data (e.g., energy demand at each location, external covariates, etc.). Failing to incorporate these features limits the model’s ability to accurately represent real-world grid dynamics, leading to less robust predictions.
    
    \item \textbf{Limited Fusion of Spatial and Temporal Information:} Many load forecasting models treat spatial and temporal information separately, either through sequential models (like LSTMs) that ignore spatial context or graph-based models that lack robust temporal modeling. This split approach fails to capture how spatial dependencies and temporal patterns interact, limiting the model's ability to adapt to sudden changes in load patterns or energy flow dynamics.
    
    \item \textbf{Lack of Integration of RES Data:} The intermittent and variable nature of RES, such as solar and wind, introduces significant uncertainty into load forecasts. Most current models do not account for RES data, leading to predictions that may not accurately reflect fluctuations in energy supply and demand.
\end{enumerate}

In this study, we present a hybrid GAT-LSTM fusion model for STLF that addresses key limitations of existing forecasting approaches. The proposed model integrates grid-specific attributes, handles spatial-temporal data fusion, and incorporates RES data to provide more realistic and reliable load predictions. By capturing complex interactions within modern power grids, our approach offers enhanced accuracy and robustness, making it well suited to the intricacies of today’s dynamic energy landscape.

The rest of the paper is organized as follows; Section~\ref{sec:review} presents reviews of key concepts, including energy forecasting techniques and graph neural networks, Section~\ref{sec:methodology} describes the data and introduces our hybrid GAT-LSTM model, Section~\ref{sec:experiments} presents the experimental results and discusses the efficacy of our approach, and finally, Section~\ref{sec:conclusion} concludes with future directions.

\section{Review of Key Concepts}
\label{sec:review}
As discussed, forecasting models predict future energy demand and supply by analyzing historical time-series data and relevant covariates to uncover important dynamics within the data. This section provides a brief overview of some widely employed energy time-series forecasting techniques in the literature.

\subsection{Traditional Energy Forecasting Models}

Traditional statistical methods for energy forecasting, such as autoregressive models and exponential smoothing techniques, are foundational approaches that use historical data to predict future energy demand. These models are widely used for linear and stationary time series data.

Autoregressive models, such as ARMA and its extensions (ARIMA and SARIMA), and exponential smoothing techniques, have been widely used for STLF. ARIMA models relationships among current values, past values, and previous errors, with SARIMA incorporating seasonal differencing for datasets with seasonality~\cite{Dannecker2015,Makridakis1997}. 
\par Exponential smoothing methods, including Holt-linear and Holt-Winters extensions, use weighted averages of past observations to capture trends and seasonality in non-stationary data~\cite{Lima2019}. These approaches are valued for their simplicity and effectiveness in linear time series data~\cite{Chodakowska2021,Lima2019,Deb2017}. However, they are limited in handling non-linear, high-dimensional data and fail to capture the complex spatial and temporal dependencies inherent in energy systems. These limitations are especially evident in modern energy systems, where demand patterns are shaped by dynamic factors such as weather, renewable energy integration, and network topology.

\subsection{Deep Learning-Based Energy Forecasting Models}

Deep learning, a subfield of machine learning, leverages neural networks to automatically learn patterns from data, making it particularly effective for modeling complex, non-linear relationships in large datasets~\cite{Berriel2017}. Sequence processing models such as recurrent neural networks (RNNs) and long-short-term memory networks (LSTMs) are widely used for time-based predictions, as they retain information across time steps and capture temporal dependencies~\cite{Werbos1990,Williams1989}.

RNNs model sequential data by updating a hidden state based on the current input and the previous hidden state. While effective for short-term dependencies, they struggle with vanishing or exploding gradients, limiting their ability to capture long-term dependencies~\cite{Colah2015,Kag2021,Mozer1991}. 

LSTMs address these challenges through a more sophisticated architecture featuring a memory cell and gating mechanisms (forget, input, and output gates) that regulate information flow. This design enables LSTMs to capture both long-term and short-term dependencies in time-series data~\cite{Gers2000, Hochreiter1997,  Gers2002}.

RNNs and LSTMs have been extensively applied to load forecasting, where they effectively model temporal dependencies and dynamic patterns in energy time-series data~\cite{Jin2022,Bianchi2017,Fang2023,PerezOrtiz2003}. However, these models are limited to temporal dependencies and do not account for the spatial interactions inherent in energy systems. To address this limitation, hybrid models such as the proposed GAT-LSTM combine the strengths of LSTMs for temporal modeling with GNNs for spatial dependencies. This integration provides a holistic representation of energy systems by capturing both time-series dynamics and grid topology.

\subsection{Graph Neural Networks (GNNs) for Energy Forecasting}

Recent advancements in deep neural networks (DNNs) have extended their applications to graph-structured (non-Euclidean) data, enabling the modeling of complex relationships inherent in graphs~\cite{Gori2005, Scarselli2008, Liao2021}. GNNs are particularly effective in capturing spatial dependencies in domains such as social networks, recommendation systems, and energy systems, where power grids naturally exhibit graph structures~\cite{Hu2024}. By combining graph topology with node and edge attributes, GNNs facilitate the modeling of spatial interactions critical for understanding the dynamics of energy networks.

\textbf{Graph-Structured Data and GNN Architecture:} A graph $G = (V, E)$ consists of nodes ($V$) and edges ($E$), which may be directed or undirected. The graph structure is represented by an adjacency matrix $A$, where $a_{ij} = 1$ if an edge exists between nodes $i$ and $j$, and $a_{ij} = 0$ otherwise. Node features $x_i$ are organized into a feature matrix $X \in \mathbb{R}^{N \times F}$, where $N$ is the number of nodes and $F$ is the number of features per node.

The core operation in GNNs is the graph convolution, where nodes aggregate information from their neighbors to update their representations. At layer $l$, this is defined as:

\begin{equation} h_i^{(l+1)} = \text{Agg}\left({h_j^{(l)} : j \in N(i) \cup {i}}\right) \label{eq:graph_convolution} \end{equation}

where $h_i^{(l)}$ is the embedding of node $i$ at layer $l$, $N(i)$ denotes the neighbors of node $i$, and $\text{Agg}$ is an aggregation function (e.g., sum, mean, or max). After aggregation, embeddings are passed through a non-linear activation function, such as ReLU.

\begin{figure}[h!]
    \centering
    \includegraphics[width=1.0\linewidth]{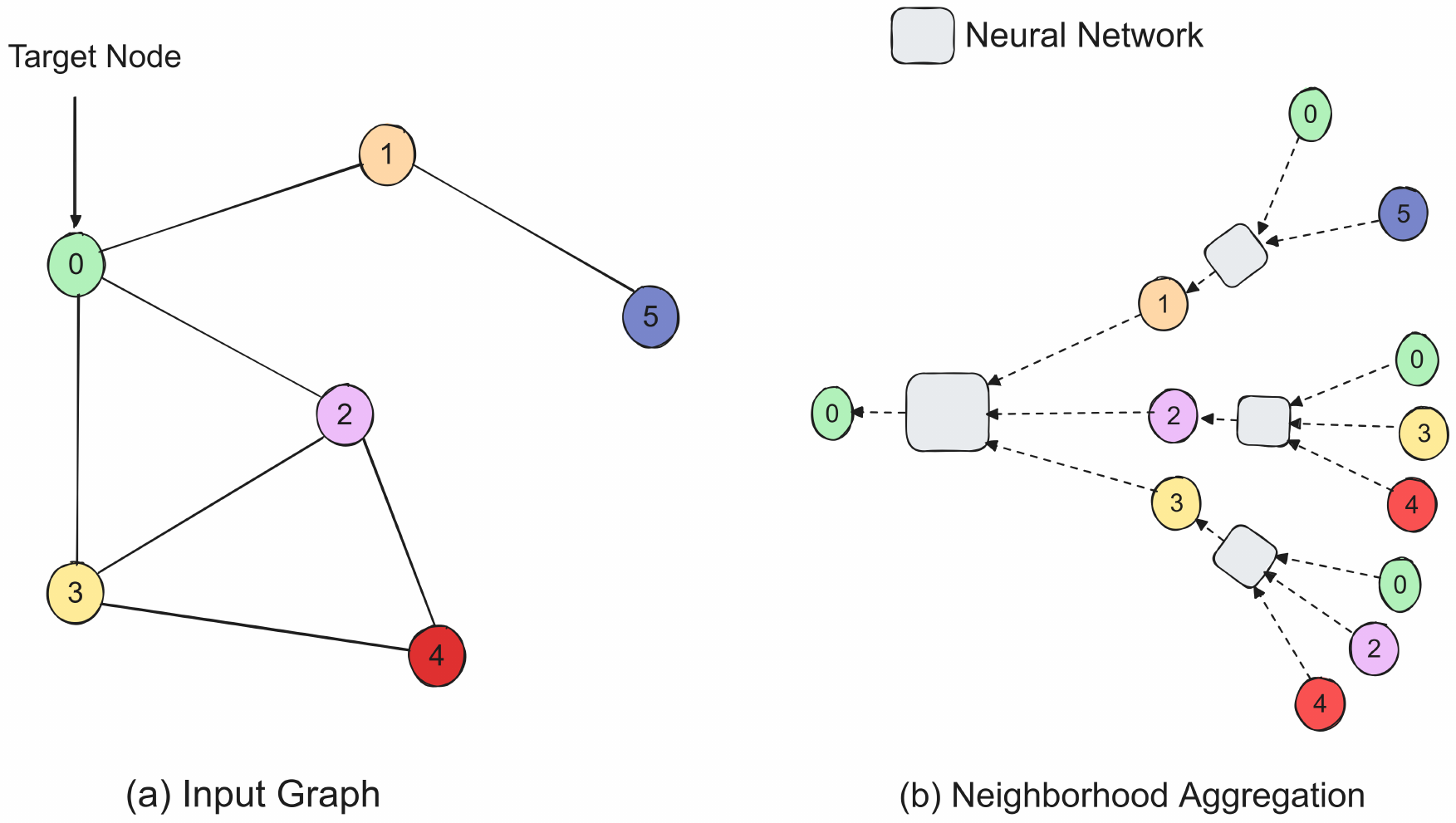}
    \caption{ Neighborhood Aggregation in GNNs. Source~\cite{Jin2022GNNLens}.}
    \label{fig:gnn_architecture}
\end{figure}

Figure~\ref{fig:gnn_architecture} illustrates neighborhood aggregation in GNNs. For energy systems, nodes may represent grid regions or substations, and edges capture relationships like shared demand or transmission lines. This process enables GNNs to learn spatial dependencies and provide context-aware predictions.

\textbf{Spectral- and Spatial-Based GNNs:} GNNs can be broadly categorized into spectral or spatial-based ~\cite{Bruna2013, Henaff2015, Chiang2019, Hamilton2017, velivckovic2017}. Spectral-based GNNs, such as Graph Convolutional Networks (GCNs), apply filters in the Fourier domain ~\cite{kipf2016,yao2019,cao2022}, leveraging the normalized adjacency matrix $\hat{A}$ to update node embeddings:

\begin{equation} H^{(l+1)} = \sigma(\hat{A} H^{(l)} W^{(l)}) \label{eq:gcn_update} \end{equation}

where $H^{(l)}$ is the feature matrix at layer $l$, $W^{(l)}$ is the weight matrix, and $\sigma$ is the activation function. GCNs are effective for tasks requiring localized information aggregation~\cite{Liao2021,Mansoor2024}.

Spatial-based GNNs, such as Graph Attention Networks (GATs), use attention mechanisms to assign importance to neighboring nodes during aggregation~\cite{bahdanau2014,gehring2016}. The attention coefficient $\alpha_{ij}$ is computed as:

\begin{equation} \alpha_{ij} = \frac{\exp(\sigma(a^T [W h_i || W h_j]))}{\sum_{k \in N(i)} \exp(\sigma(a^T [W h_i || W h_k]))} \label{eq:gat_attention} \end{equation}

where $a$ is a learnable attention vector and $||$ denotes concatenation. Multi-head attention refines this process further, enhancing robustness~\cite{huang2023}.

\textbf{Applications and Limitations in Load Forecasting:} GNNs, including GCNs and GATs, have been used to improve load forecasting by modeling grid topology and spatial dependencies~\cite{Mansoor2024, Liao2021,huang2023}. However, GNNs alone cannot capture temporal dependencies inherent in energy systems, which are critical for forecasting dynamic load patterns.

\subsection{Hybrid Models for Energy Forecasting}

Hybrid models, such as the proposed GAT-LSTM, address the limitations of standalone GNNs and LSTMs by combining their strengths. While GATs model spatial dependencies in grid topology, LSTMs handle temporal dynamics by capturing sequential patterns in energy consumption. This integration enables a holistic representation of energy systems, accounting for both spatial interactions and time-series dynamics~\cite{Mansoor2024,Eandi2022, Lin2021, huang2023}.

The novelty of the GAT-LSTM lies in its ability to adapt graph-structured relationships and temporal patterns simultaneously. For example, GATs use attention mechanisms to focus on the most relevant grid regions, while LSTMs capture the evolving energy consumption trends. Together, they synergize to improve forecasting accuracy, particularly for complex, interconnected energy networks where traditional models and standalone deep learning methods fall short.

This hybrid approach has been extensively studied in domains like traffic forecasting~\cite{wu2018,zhu2024,zhang2020}, but remains underexplored in energy systems. By incorporating grid-specific features and leveraging the spatio-temporal interplay, the GAT-LSTM provides robust and context-aware predictions for energy time series forecasting, marking a significant advancement in the field.

\section{Methodology}
\label{sec:methodology} 
\subsection{Data}

\subsubsection{\textbf{Data Source and Description}:}

For this study, we used the Brazilian power system as a case study with actual data on various aspects of the power system and covariate factors. The data sources and description is given in Tables~\ref{tab:data_sources} and~\ref{tab:data_description}, respectively. Except for the graph data, which are static, all other aspects of the dataset represent 2-years worth of data from 2019--2020.

\begin{table}[h!]
  \caption{Data Sources}
  \label{tab:data_sources}
  \begin{tabular}{lcl}
    \toprule
    Data & Details & Source \\
    \midrule
    Electricity & Load, PV, wind, etc. & \cite{deng2023harmonized} \\
    Grid & Line length, capacity, efficiency, etc. & \cite{deng2023harmonized} \\
    Weather & Temperature, pressure, rainfall, etc. & \cite{BrazilWeather} \\
    Socio-economic & State-wise GDP & \cite{BrazilGDP} \\
    Population & State-wise population & \cite{BrazilPopulation} \\
    \bottomrule
  \end{tabular}
\end{table}

\begin{table}[h!]
  \caption{Data Description}
  \label{tab:data_description}
  \begin{tabular}{lp{4.3cm}l}
    \toprule
    \textbf{Category} & \textbf{Details} & \textbf{Horizon} \\
    \midrule
    \textbf{Sequence Data} & Atmospheric pressure, Total hourly rain, Global radiation, Air temperature, Dew point temperature, Relative humidity, Wind direction, Wind maximum gust, Wind speed, PV generation, Onshore wind generation, Offshore wind generation, Load profile (by consumption) & 1-hour \\
    \midrule
    \textbf{Graph Data} &
    \textbf{Node features:} Source (state), Target (state), PV potential, Onshore wind potential, Offshore wind potential, geometry (longitude \& latitude) \newline
    \textbf{Edge attributes:} Line capacity, Line efficiency, Line length, Line carrier & Static \\
    \midrule
    \textbf{Socio-economic} & Population, GDP, Total plant capacity & Annual \\
    \midrule
    \textbf{Calendar} & Year, Quarter, Month, Day, Hour, Day-of-week, Week-of-Year, Holiday, Season & Annual \\
    \bottomrule
  \end{tabular}
\end{table}

\subsubsection{\textbf{Data Preprocessing}:}

The data were cleaned and preprocessed for modeling through the following steps:
\begin{enumerate}
    \item \textbf{Consolidation of Weather Data:} The weather data consisted of observations from various stations in different states of Brazil, with some states having multiple stations. To ensure consistent hourly data for each weather variable per state, we calculated the mean and standard deviation (STD) of each variable, grouped by state and datetime.

    \item \textbf{Missing Value Imputation:} Analysis showed approximately 21.42\% of missing data across the weather dataset, which was filled by interpolation of time series to maintain continuity in the data.

    \item \textbf{Handling Insufficient Data:} One state had only a single weather station, resulting in NaNs for the STD values. We addressed this by setting the STD values for that state to zero.

    \item \textbf{Negative Value Correction:} Negative values were observed in the PV variable, which were corrected by replacing them with zero.

    \item \textbf{Data Scaling:} To handle variance and outliers, we scaled the dataset using RobustScaler from the Python Scikit-learn package.

    \item \textbf{Data Splitting:} A dynamic split was performed to create training, validation, and test sets as follows:
    \begin{itemize}
        \item Training set: January-December 2019
        \item Validation set: January–June 2020
        \item Test set: July–December 2020
    \end{itemize}

    \item \textbf{Target Variable Creation:} Target values for prediction were created by shifting the current load by one hour to facilitate next-hour load forecasting.

    \item \textbf{State-wise Sequencing:} Finally, we generated state-wise sequences, aligning data by state and timestamp to feed into the model.
\end{enumerate}

\subsection{Problem Definition and Model Architecture}

Let $X = \{x_1, x_2, \dots, x_p\}$ represent a historical sequence of multidimensional variables in $p$ time steps, where each $x_t \in \mathbb{R}^N$ is a vector of $N$ features at time $t$. Each vector $x_t$ includes the power load at time $t$ and other covariates that influence future load (e.g., temperature, seasonal and holiday variables, etc.). The objective of this study is to forecast the next hour's power load, $y_{p+1}$, by leveraging these historical data alongside the spatial grid information and RES data.

\begin{figure*}[h!]
    \centering
    \includegraphics[width=\textwidth]{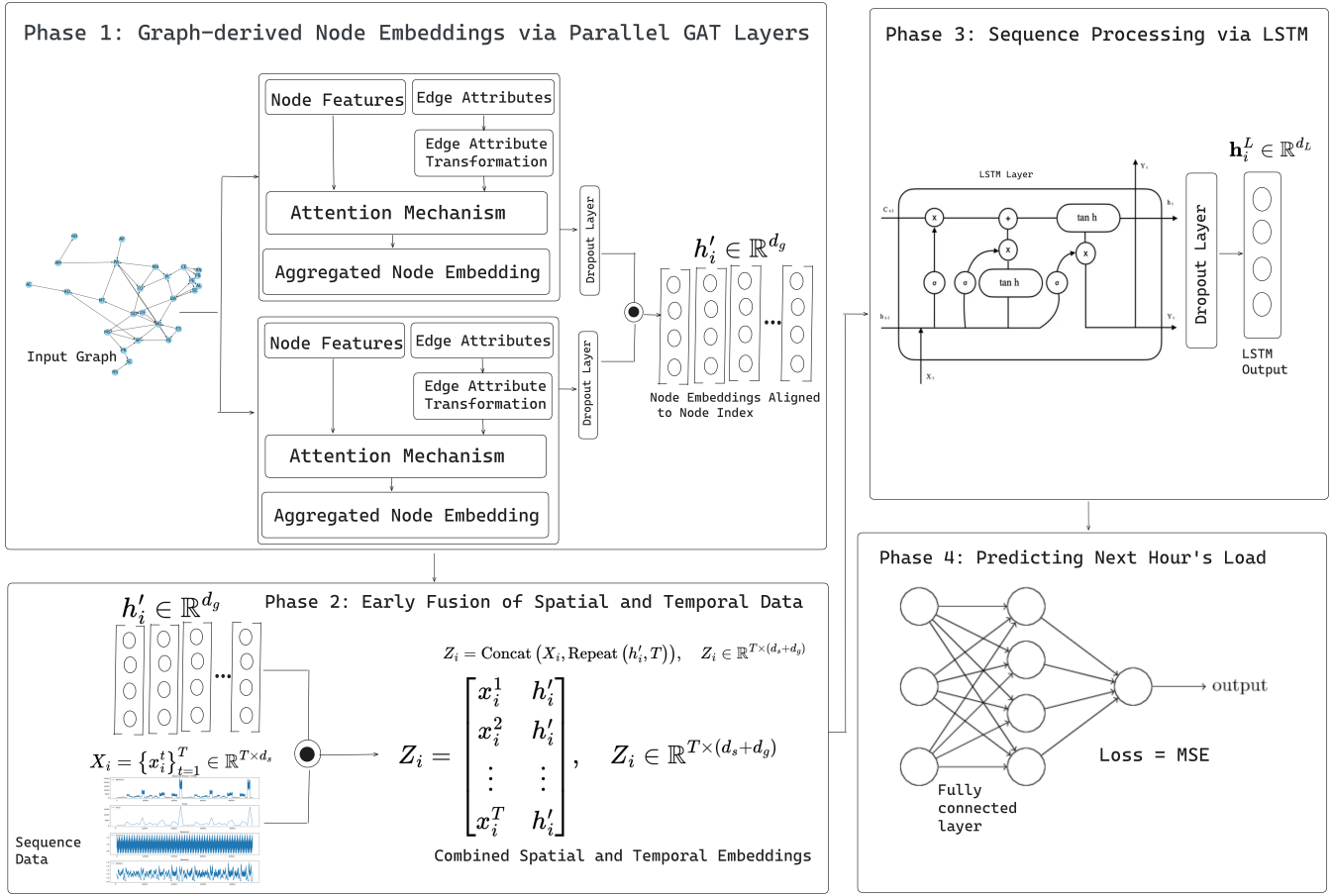}
    \caption{Model Architecture}
    \label{fig:Model architecture}
\end{figure*}

To achieve this, we propose a hybrid GAT-LSTM model \footnote{Our model code is available under an open license: \url{https://github.com/ugoorji12/Load-Forecasting-using-GAT-LSTM}} as shown in Figure~\ref{fig:Model architecture} and described below:
\subsubsection*{\textbf{Phase 1 (Get Node Embeddings):}} The model employs 2 parallel GAT layers to compute node embeddings that reflect spatial dependencies and grid-specific constraints. Unlike traditional attention mechanisms, which typically rely only on node features, our approach also incorporates edge attributes (i.e., line capacities, efficiencies, etc.) into the calculation of attention coefficients. Thus, we modify Equation~\ref{eq:gat_attention} to incorporate edge attributes such that, for a pair of connected nodes $i$ and $j$ with respective node features $h_i$ and $h_j$ and edge attribute $e_{ij}$, the attention coefficient $\alpha_{ij}$ becomes:

\begin{equation}
\alpha_{ij}^k = \frac{\exp\left(\sigma\left(a^k \cdot [W^k h_i^{'} \parallel W^k h_j^{'} \parallel U^k e_{ij}^{'}] \right)\right)}{\sum_{k \in N(i)} \exp\left(\sigma\left(a^k \cdot [W^k h_i^{'} \parallel W^k h_k^{'} \parallel U^k e_{ik}^{'}]\right)\right)} 
\label{eq:attention_edge}
\end{equation}

Where:
\begin{itemize}
    \item $W^k$ = learnable transformation matrix specific to attention head $k$ of node features $h_i^{'}$
    \item $U^k$ = learnable transformation matrix specific to attention head $k$ of edge attributes $e_{ij}^{'}$
    \item $a^k$ = learnable vector that projects the concatenated inputs into a scalar (used for attention scoring)
    \item $\sigma$ = LeakyReLU activation function
\end{itemize}

Based on the multi-head attention mechanism, the aggregation and update function for each head is given as:

\begin{equation}
h_i^{'} = \text{Concat}\left(\sigma\left(\sum_{j \in N(i)} \alpha_{ij}^k W^k h_j^{'}\right), \forall k \in \{1, \dots, K\}\right)
\label{eq:multi_head_attention}
\end{equation}

This formulation allows the model to dynamically learn the importance of each connection based on both node and edge attributes, yielding node embeddings that better capture the underlying power grid structure and constraints.
\subsubsection*{\textbf{Phase 2 (Early Fusion of Spatial and Temporal Data):}} The goal of this phase is to combine the graph-derived node embeddings $h_i^{'}$ from the GAT layer (Phase 1) with temporal sequence data. We achieve this by first expanding the $h_i^{'}$ along the temporal dimension to match the sequence data as represented in Equation~\ref{eq:expand_embeddings}.

\begin{equation}
Z_i = 
\begin{bmatrix}
x_i^1 & h_i^{'} \\
x_i^2 & h_i^{'} \\
\vdots & \vdots \\
x_i^T & h_i^{'}
\end{bmatrix}, \, Z_i \in \mathbb{R}^{T \times (d_s + d_g)}
\label{eq:expand_embeddings}
\end{equation}

The final concatenation is given as:

\begin{equation}
Z_i = \text{Concat}(X_i, \text{Repeat}(h_i^{'}, T)), \, Z_i \in \mathbb{R}^{T \times (d_s + d_g)}
\label{eq:concat_embeddings}
\end{equation}

Where:
\begin{itemize}
    \item $h_i^{'} \in \mathbb{R}^{d_g}$ = The graph-derived embedding for node $i$, where $d_g$ is the GAT output dimension (after concatenation across heads).
    \item $X_i = \{x_i^t\}_{t=1}^T \in \mathbb{R}^{T \times d_s}$ = The temporal sequence data for node $i$, $T$ is the sequence length (e.g., 24 hours), and $d_s$ is the feature dimension of the sequence data.
\end{itemize}

This early fusion strategy ensures that both spatial and temporal information are jointly modeled in the downstream LSTM layer.

\subsubsection*{\textbf{Phase 3 (LSTM Layer for Sequential Processing):}} The combined spatial-temporal data are fed into an LSTM layer, which learns the combined dependencies, capturing how past load trends and covariates evolve over time to influence future loads. This layer enables the model to retain and process the long-term spatio-temporal dependencies in the data.

\subsubsection*{\textbf{Phase 4 (Final Prediction Layer):}} The output from the LSTM layer is fed into a fully connected layer that generates the forecast for the next hour's load, $y_{p+1}$.

\subsection{Training Process}
\subsubsection*{\textbf{Loss Function and Optimization:}} The training process used the mean squared error (MSE) as the loss function for both training and validation. MSE is particularly suited for regression tasks as it heavily penalizes larger errors, ensuring a focus on minimizing significant deviations in predictions.

For optimization, the Adam optimizer was chosen due to its effective combination of momentum and adaptive learning rates, which makes it well suited for handling complex models. Additionally, a \texttt{ReduceLROnPlateau} learning rate scheduler was used to dynamically adjust the learning rate. If the validation loss did not improve for five consecutive epochs, the learning rate was reduced by a factor of 0.1 (90\% decrease), promoting more effective convergence.

To prevent overfitting, Early Stopping was applied, halting the training process if validation loss showed no improvement for 10 consecutive epochs. This ensured that the training process was stopped once convergence was achieved.

These techniques work together to ensure efficient learning, smoother convergence, and better generalization.

\section{Experiments and Results}
\label{sec:experiments} 
\subsection{Experimental Setup}

The experiments were carried out in a high performance computing environment featuring a dual-socket architecture with 12 physical cores (24 logical CPUs) and 192 GB of memory. The system includes 3 NVIDIA Tesla GPUs and ran on Linux (Debian 6.1), with Slurm (version 22.05.8) used for job scheduling. The hyperparameters used in the model were carefully selected, with details provided in Table~\ref{tab:model_parameters}.

\begin{table}[h!]
  \caption{Description of Model Parameters}
  \label{tab:model_parameters}
  \centering
  \begin{tabular}{lc}
    \toprule
    \textbf{Parameter} & \textbf{Value} \\
    \midrule
    Sequence length & 24 \\
    Batch size & 27 \\
    GAT output layer & 64 \\
    GAT attention heads & 8 \\
    LSTM hidden-state & 128 \\
    Number of LSTM layers & 4 \\
    Learning rate & 0.0001 \\
    Weight decay & $1 \times 10^{-5}$ \\
    GAT Dropout & 0.2 \\
    LSTM Dropout & 0.3 \\
    Epochs & 200 \\
    \bottomrule
  \end{tabular}
\end{table}

\subsection{Baseline Models}

The following baseline models were used to compare the performance of our proposed approach:

\textbf{GCN-LSTM:} This hybrid model combines Graph Convolutional Network (GCN) layers with a Long Short-Term Memory (LSTM) network for load forecasting. It mirrors the GAT-LSTM architecture by applying two parallel 1-hop GCN layers, each performing a 1-hop convolution on the node features, followed by dropout for regularization. The outputs of these GCN layers are concatenated and indexed for the nodes corresponding to the input sequences. These combined GCN features are expanded to match the temporal sequence length and concatenated with the sequence data. The LSTM layer captures temporal dependencies, and a fully connected layer generates the final output. Unlike GAT-LSTM, this model does not incorporate edge attributes, as standard GCNs focus exclusively on node features and graph structure.

\textbf{EdgeGCN-LSTM:} An extension of GCN-LSTM, this model integrates edge attributes into the message-passing process using a custom \texttt{EdgeAttrGCNConv} layer. Edge attributes are transformed via a linear layer before aggregation, allowing the model to utilize both node and edge information effectively. The aggregated features are passed to an LSTM layer to capture temporal dynamics and generate predictions.

\textbf{LSTM:} Serving as a sequence forecasting baseline, this model focuses solely on temporal data without incorporating graph-based features. It uses an LSTM network to learn temporal dependencies and produce forecasts based on dynamic time-series data.

\textbf{XGBoost:} A popular tree-based regression model, XGBoost is used here as a baseline for evaluating performance on time-series data. Like the LSTM model, it does not include graph-based features and relies entirely on dynamic data for forecasting.

\subsection{Evaluation Metrics}

Our evaluation metrics include Mean Absolute Error (MAE), Root Mean Square Error (RMSE), and Mean Absolute Percentage Error (MAPE).

Concretely, these metrics are widely used to evaluate the accuracy of single or hybrid load forecasting models~\cite{Pinson2013,Mamun2020}.

\subsection{Experiment Results and Discussion}

\textbf{Training Performance and Early Stopping Analysis}

\begin{figure}[h!]
    \centering
    \includegraphics[width=1.0\linewidth]{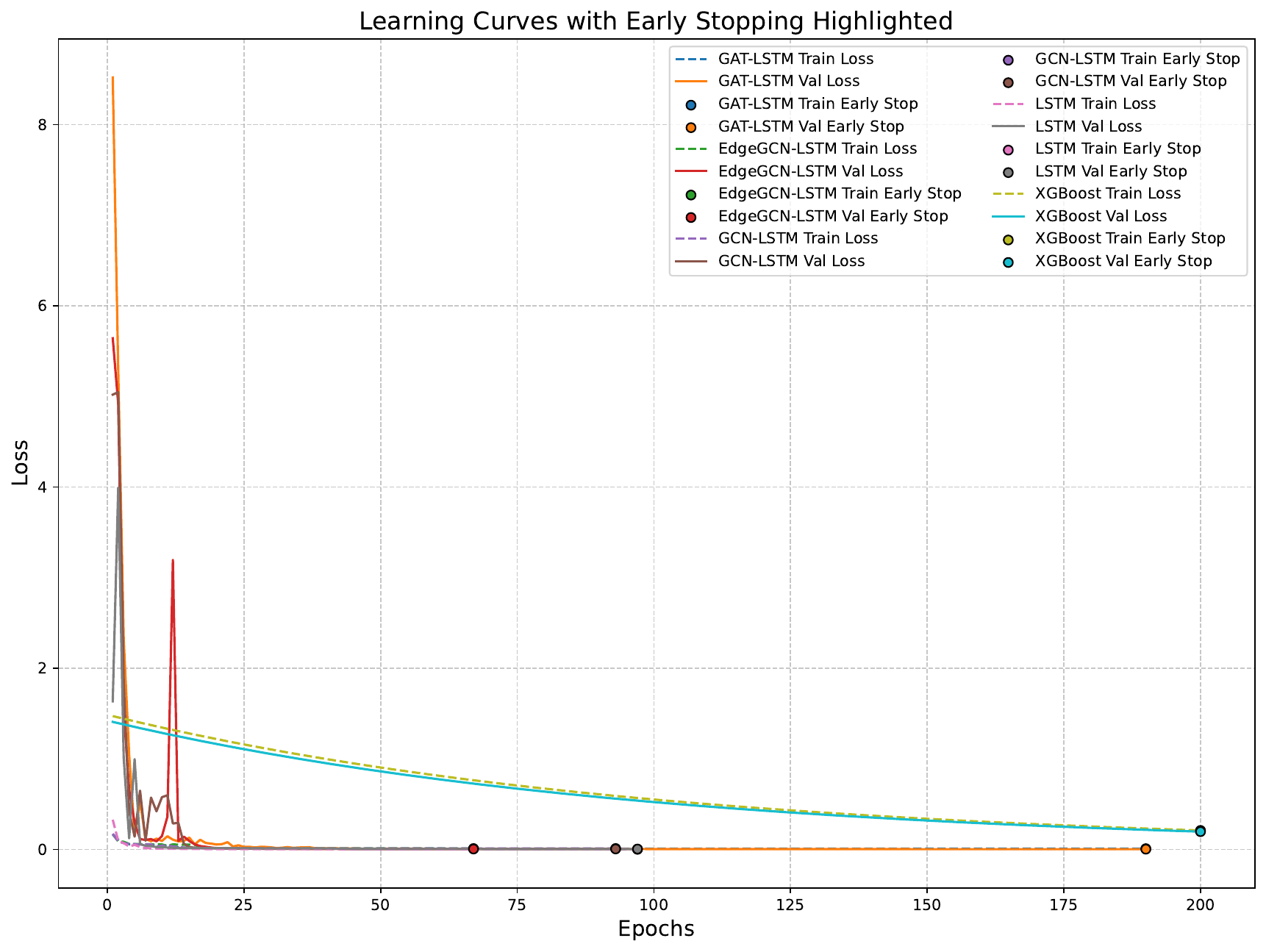} 
    \caption{Learning curve for all models.}
    \label{fig:learning_curves}
\end{figure}

Figure~\ref{fig:learning_curves} presents the learning curves for all models during training, highlighting their convergence behavior and potential overfitting risks. GAT-LSTM demonstrates rapid convergence with stable training and validation losses, suggesting efficient training and generalization. Early stopping was applied around optimal epochs to prevent overfitting. In contrast, EdgeGCN-LSTM and GCN-LSTM show steady declines in training loss but exhibit fluctuations in validation loss, indicating minor risks of overfitting. LSTM converges more slowly, with its validation loss stabilizing at a higher value, reflecting its weaker capacity to model the data comprehensively. Lastly, XGBoost displays the poorest training performance, with consistently high errors throughout, underscoring its limitations in capturing temporal dependencies.

\textbf{Model Comparison Based on Accuracy Metrics}

\begin{table}[h!]
  \caption{Experiment Results}
  \label{tab:experiment_results}
  \centering
  \resizebox{\columnwidth}{!}{%
  \begin{tabular}{lccc}
    \toprule
    \textbf{Model} & \textbf{MAE (MW)} & \textbf{RMSE (MW)} & \textbf{MAPE (\%)} \\
    \midrule
    \textbf{GAT-LSTM} & \textbf{64.64} & \textbf{119.06} & \textbf{4.59} \\
    LSTM & 82.68 & 141.55 & 5.75 \\
    EdgeGCN-LSTM & 84.63 & 148.09 & 7.24 \\
    GCN-LSTM & 89.11 & 184.12 & 5.72 \\
    XGBoost & 297.47 & 517.69 & 40.50 \\
    \bottomrule
  \end{tabular}%
  }
\end{table}

Table~\ref{tab:experiment_results} summarizes the performance of all models based on MAE, RMSE, and MAPE. GAT-LSTM achieves the best overall accuracy, outperforming other models across all metrics. Specifically, it shows a 21. 82\% improvement in MAE compared to LSTM and a 23.62\% improvement over EdgeGCN-LSTM. This highlights the ability of GAT-LSTM to effectively capture both spatial and temporal dependencies, leveraging the graph-attention mechanism. LSTM, despite lacking spatial awareness, performs better than both EdgeGCN-LSTM and GCN-LSTM due to its robust temporal modeling capabilities. The underperformance of EdgeGCN-LSTM relative to LSTM suggests that incorporating spatial relationships without attention mechanisms might introduce irrelevant or noisy information, hindering forecasting accuracy. Similarly, GCN-LSTM performs the worst among GNN-based models, indicating that suboptimal spatial features can reduce model effectiveness. Lastly, XGBoost demonstrates the highest errors across all metrics, reflecting its inability to model non-linear temporal dependencies and spatial relationships critical for load forecasting.

\textbf{Analysis of Mean Actual vs Predicted Load Curves}

\begin{figure}[h!]
    \centering
    \includegraphics[width=1.0\linewidth]{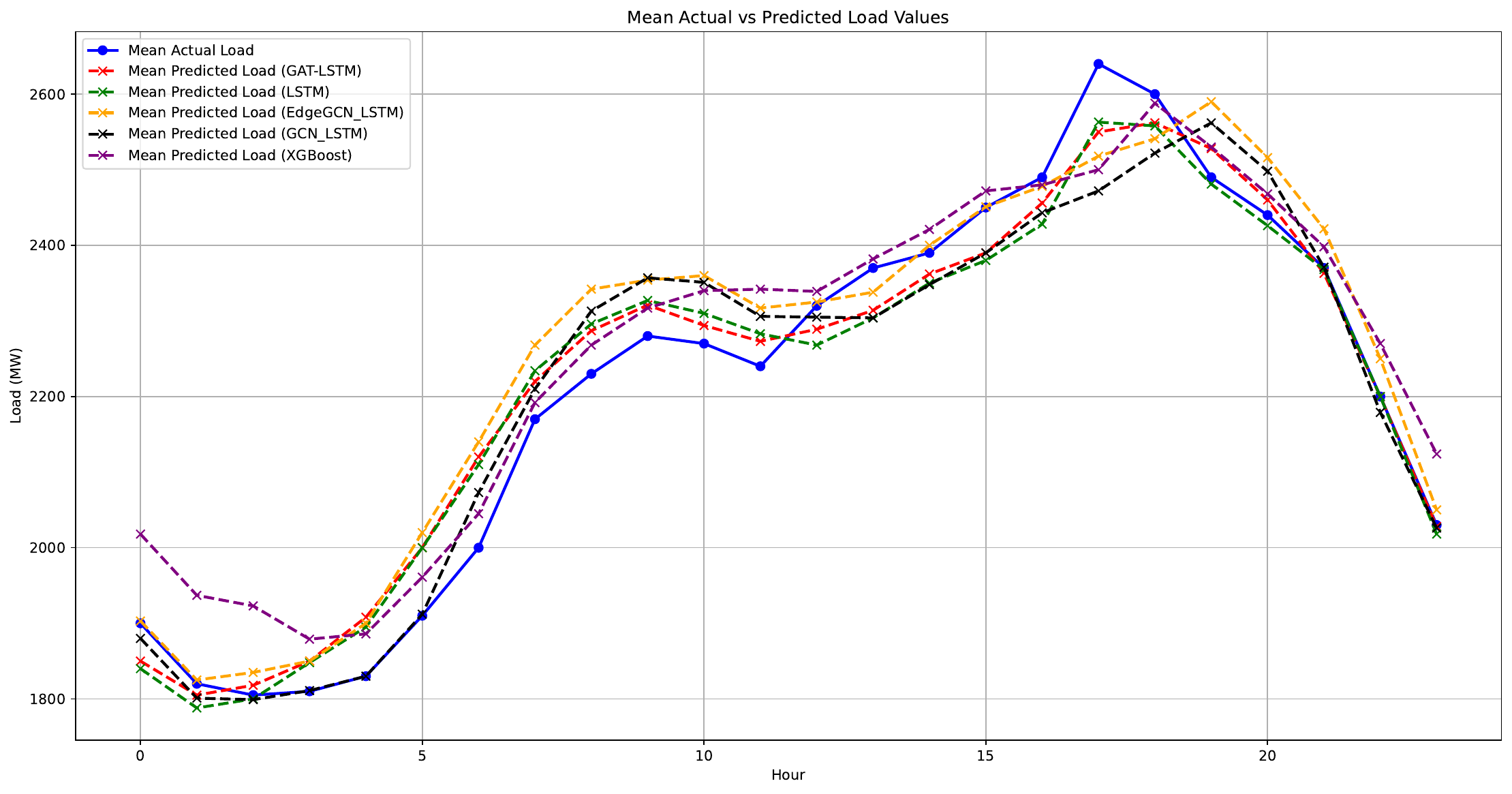} 
    \caption{Mean Actual vs Predicted Load Values for all Models.}
    \label{fig:mean_actual_vs_predicted}
\end{figure}

Figure~\ref{fig:mean_actual_vs_predicted} compares the mean actual and predicted load profiles across all models. GAT-LSTM closely follows the actual load curve throughout the day, reflecting its superior accuracy and ability to generalize across varying load conditions. LSTM and EdgeGCN-LSTM show reasonable alignment with the actual load curve but exhibit larger deviations during peak and off-peak transitions, particularly in high-gradient periods. GCN-LSTM struggles more during peak hours, with noticeable deviations during high-load periods. In contrast, XGBoost displays significant errors, including peak and off-peak times, confirming its limited ability to generalize load dynamics effectively.

\textbf{Performance of GAT-LSTM During Peak vs Off-peak Hours}

\begin{figure}[h!]
    \centering
    \includegraphics[width=1.1\linewidth]{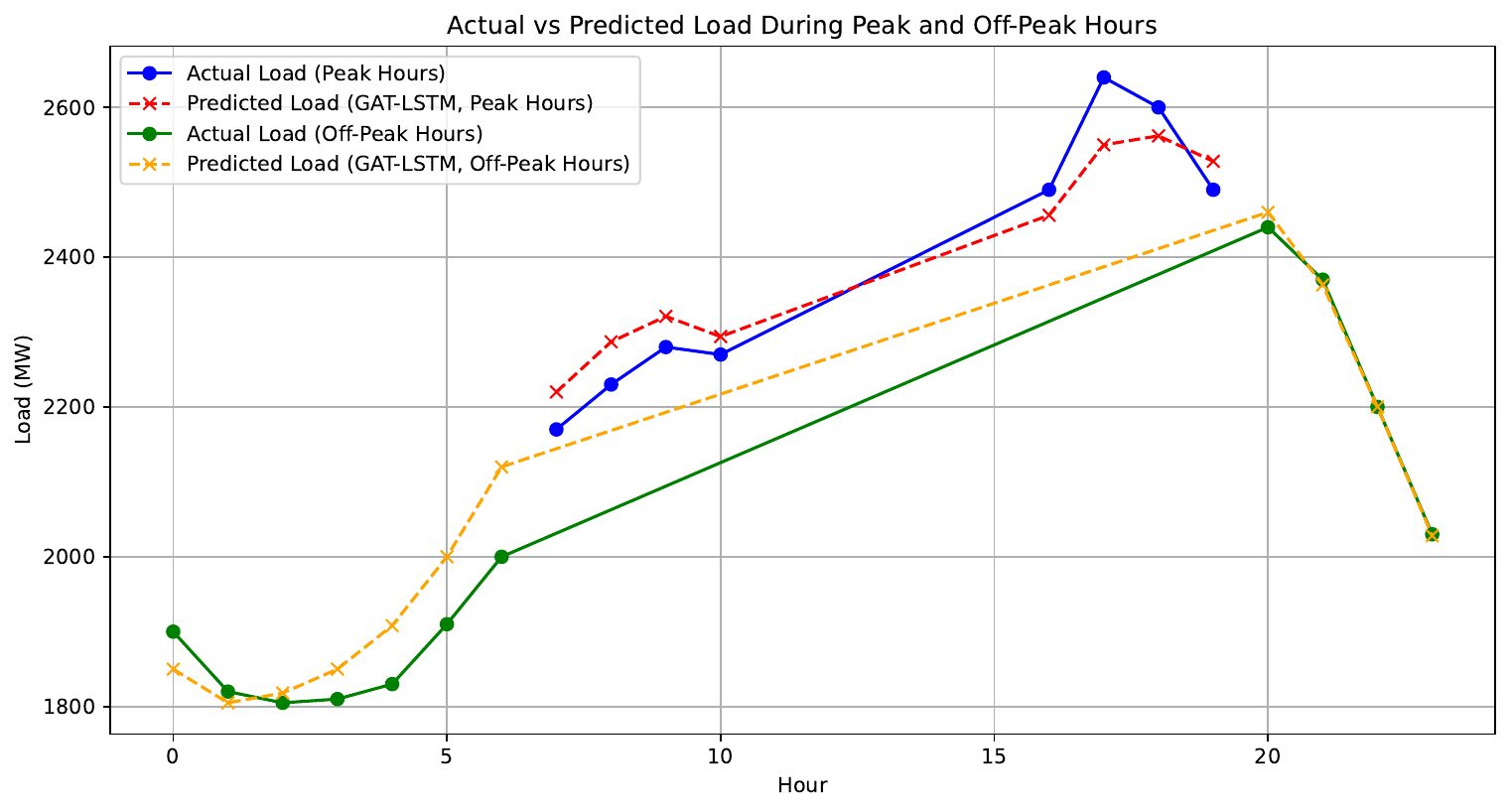} 
    \caption{Peak vs Off-peak Performance for GAT-LSTM.}
    \label{fig:peak vs off_peak}
\end{figure}

Figure~\ref{fig:peak vs off_peak} illustrates the average performance of GAT-LSTM during peak and off-peak periods across the test set. During peak hours (7:00 AM–10:00 AM and 4:00 PM–7:00 PM), GAT-LSTM consistently tracks actual load trends but exhibits slight deviations during sharp increases, such as the morning ramp-up. Evening peaks are generally better captured, although minor overestimations occur after the peak (e.g., hours 18–19). During off-peak hours (12:00 AM–6:00 AM and 8:00 PM–11:00 PM), predictions align well with actual loads, although a pattern of overestimation is observed during early off-peak periods (e.g., hours 2–5). This performance demonstrates the model's robustness, but also highlights areas for further improvement, such as better handling of high-gradient patterns.

\subsection{Key Insights and Discussion}

The experimental results underscore several critical insights:

\textbf{Effectiveness of Attention Mechanisms:} The GAT-LSTM model demonstrates superior accuracy by effectively capturing spatial and temporal dependencies using attention mechanisms. Its ability to prioritize relevant node and edge attributes enables robust modeling of grid-specific constraints, such as line capacities. However, the model struggles slightly during rapid transitions, such as morning ramp-ups, suggesting that the temporal component could benefit from further fine-tuning. The integration of spatial embeddings and temporal features remains a key strength, enabling GAT-LSTM to model complex interactions with high robustness for real-world electricity forecasting.

\textbf{Limitations of Basic Graph Architectures:} Both EdgeGCN-LSTM and GCN-LSTM perform worse than LSTM, indicating that spatial features, when modeled without attention mechanisms, may introduce noise or irrelevant information, reducing forecasting accuracy.

\textbf{Temporal Strength of LSTM:} Despite its lack of spatial awareness, LSTM's strong temporal modeling capabilities allow it to outperform EdgeGCN-LSTM and GCN-LSTM. This highlights the importance of robust temporal modeling in energy forecasting tasks.

\textbf{Limitations of XGBoost: }XGBoost exhibits the highest errors, confirming its inability to model sequential relationships and temporal dependencies effectively. Its tree-based approach further limits its ability to capture the non-linear temporal dynamics critical for load forecasting.

\section{Conclusion, Limitations, and Future Work}
\label{sec:conclusion}
In this paper, we introduce and evaluate the GAT-LSTM model for hourly power load forecasting, combining GAT and LSTM to effectively capture spatial and temporal dependencies in electricity grids. Our results demonstrate that GAT-LSTM consistently outperforms state-of-the-art models, across key metrics such as MAE, RMSE, and MAPE. This superior performance arises from the model's ability to leverage graph-based attention mechanisms to extract meaningful spatial features while utilizing the LSTM’s strength in modeling sequential patterns.
\par Despite its strong performance, the model has notable limitations. The integration of graph attention mechanisms and LSTM layers introduces significant computational complexity, potentially hindering scalability for large datasets or resource-constrained environments. Furthermore, the model's accuracy is influenced by the quality of the graph structure; incomplete or inaccurate node connections, often due to missing or imprecise data in electrical grid representations, can reduce its effectiveness. These issues are relatively common in real-world grids due to challenges such as missing data, approximations in topology, and outdated infrastructure records. Additionally, while GAT-LSTM performs well during stable off-peak periods, it struggles to fully capture rapid transitions, such as morning ramp-ups, leading to deviations from actual load values. This limitation is not unique to GAT-LSTM, but reflects a broader challenge in forecasting highly dynamic events, where past information may not adequately represent future behavior.
\par Future work should address these challenges by enhancing the model’s ability to handle rapid load transitions during peak hours and reducing sensitivity to low-magnitude variations in off-peak periods. Adaptive learning techniques, automated graph refinement, and the inclusion of additional external covariates, such as market data and maintenance records, could improve the model's robustness and contextual understanding. Leveraging attention mechanisms to identify and prioritize critical graph elements, such as influential nodes or edges, can further enhance interpretability and provide deeper insights into the spatial-temporal factors driving predictions. Additionally, integrating uncertainty quantification methods alongside these mechanisms would improve decision-making reliability and broaden the model's applicability in real-world scenarios.
\par In conclusion, the GAT-LSTM model represents a significant advancement in power load forecasting by effectively modeling spatial-temporal relationships in electricity grids. Its ability to identify and prioritize informative graph components allows it to capture critical spatial dependencies that underpin accurate predictions. Addressing its limitations related to computational complexity and peak-hour accuracy, along with further refinements for scalability and reliability, will enhance its practicality. Visualizing the graph structure to identify central or influential nodes could also provide valuable insights, improving both interpretability and optimization. These advancements will solidify GAT-LSTM's role as a powerful tool for grid management, demand response, and energy planning in dynamic and interconnected energy systems.

\begin{acks}
\small{This research is part of the project Innovation Lab for Utilities on Sustainable Technology and Renewable Energy (ILUSTRE), No.KICH3.LTP.20.006 of the research program LTP ROBUST which is partly financed by the Dutch Research Council (NWO)}.
\end{acks}

\bibliographystyle{ACM-Reference-Format}
\bibliography{GAT-LSTM_paper}

\end{document}